\documentclass[a4paper]{article}
\usepackage{graphicx}
\usepackage{twocolceurws}
\usepackage{lipsum}
\usepackage{hyperref}
\usepackage{url}

\title{Cost Analysis of Human-corrected Transcription for Predominately Oral Languages}

\author{
Yacouba Diarra \\ diarray
\and
Nouhoum Coulibaly  \\ n.coulibaly
\and
Michael Leventhal \\ mleventhal
}

\institution{RobotsMali AI4D Lab | @robotsmali.org}

\begin{document}
\maketitle

\begin{abstract}
Creating speech datasets for low-resource languages is a critical yet poorly understood challenge, particularly regarding the actual cost in human labor. This paper investigates the time and complexity required to produce high-quality annotated speech data for a subset of low-resource languages, low literacy Predominately Oral Languages, focusing on Bambara, a Manding language of Mali. Through a one-month field study involving ten transcribers with native proficiency, we analyze the correction of ASR-generated transcriptions of 53 hours of Bambara voice data. We report that it takes, on average, 30 hours of human labor to accurately transcribe one hour of speech data  under laboratory conditions and 36 hours under field conditions. The study provides a baseline and practical insights for a large class of languages with comparable profiles undertaking the creation of NLP resources.
\end{abstract}

\section{Introduction}

\label{section:1}
In 2020, Joshi et al defined a low-resource language as one for which large amounts of textual data needed to train NLP systems are not available and estimated that of the world’s 7,000+ languages, only 20 are well-resourced and only 100 have NLP resources \cite{joshi2020}. In 2022, Adebara and Abdul-Mageed estimated, for African languages, that only 10 have moderate resources \cite{adebara2022}. Roughly 3,000 languages have an established writing system \cite{ethnologue2023}, a necessary prerequisite for developing conventional NLP resources. We use the term \textbf{Predominately Oral Language} (POL) to describe these languages: languages where speech, to the near but not complete exclusion of writing, has been the dominant means of transmission of knowledge. The designation low-resource language is too broad to allow useful characterization of the challenges to the creation of NLP resources presented by POL as a group. We propose a subdivision of POL into three categories using literacy as the classification factor:

\begin{enumerate}
    \item \textbf{Moderate Literacy}: There exists within the language community a tradition of writing, there is adult literature, and there is a significant body of readers and writers in the language. African languages that fall into this category include a few languages with hundreds of years of writing tradition (Swahili, Amharic, Hausa) and others that have developed a writing culture in the 20th century (Yoruba, Zulu, Setswana, Gikuyu, Igbo) \cite{irele2004}.
    \item \textbf{Low Literacy}: There exists a small body of written texts such as translations, children’s books, and official documents, there is a relatively stable orthography, and there is some measure of institutional support, but there are only a small number of people who can read the language fluently and a handful capable of robust written expression. African languages that fall into this category include Shona \cite{magwa2002}, Ndebele \cite{hadebe2002}, Lingala \cite{meeuwis2020}, Bambara \cite{konta2014}, Wolof \cite{evers2017}, and Oromo \cite{bijiga2015}.
    \item \textbf{Negligible Literacy}: These languages may have no organic written texts and, possibly, a handful of translations such as excerpts of the Bible, no standardized script or orthography, no writers, and either no readers or a handful of people capable of reading a venerated text such as the Bible. As only a small number of POL fall into the first two categories, this is, necessarily, the category for the vast majority of African languages with writing systems.
\end{enumerate}

This study of Bambara considers the situation of languages in the second category of \textit{low literacy} POL where a major bottleneck to the creation of NLP resources is the scarcity of people capable of reading and writing in the language. This makes the development of speech technologies for these languages a task constrained by human resources.

Most speech recognition models need several hundred hours of training data to achieve performance sufficient for practical application. Investments in AI for development are now supporting multiple language initiatives in Africa and beyond\footnote{A collaborative of development funders recently announced an investment of over USD 100 million for AI for development, covering not just language AI but also other relevant applications, see https://www.ai4d-collaborative.org/}. While there are now limited financial resources to support African NLP initiatives, the need for a good understanding of the work necessary to develop functional speech technology for target languages leads to the research question of this study: \textbf{\textit{How much does it actually cost to create NLP resources for POLs?}}

Synthetic data generation is often the go-to option for many researchers when it comes to POLs. Building on the recent progress of Large Language Models in generating text in many African languages and advances in TTS systems, DeRenzi et al. investigated the creation of synthetic text and voice data for 10 African languages and they argue that it is feasible to create high-quality synthetic voice data for some African languages, specifically Hausa, Dholuo, and Chichewa for less than 1\% the cost of human data \cite{derenzi2025syntheticvoicedataautomatic}. 

Nevertheless, synthetic data, by nature, often exhibits an unnatural or highly patterned structure reflective of the Language Model's training distribution rather than real-world linguistic variation, which make models struggle when encountering genuine human speech or text. Relying heavily on these artificial patterns risks creating systems that are effective only within their synthetic domain, resulting in poor generalizability and the propagation of inherent biases present in the generating distribution \cite{moslem2024leveragingsyntheticaudiodata}. While we might observe improvements when used for data augmentation, the value of synthesized speech as ASR training data remains dramatically less than that of real speech \cite{Rosenberg2019SpeechRW}.

Another viable approach to creating high quality speech data that completely eliminates the need for transcription consists of creating READ datasets where native speakers are recorded reading pre-crafted sentences \cite{ardila-etal-2020-common}, \cite{panayotov2015librispeech}. In that framework, the heavy load shifts from transcription to engineering read sentences to produce discourse that approximates natural speech with linguistic variety. It assumes an accessible pool of speakers with at least reading skills in the language - that is, the POL must fall into the \textit{moderate literacy} category. In an attempt to make this approach practical for Igbo, Yoruba and Hausa, Emezue et al revisited the concept by adding a human agent in the pipeline, the so called \textit{facilitators} were in charge of handling the data collection app, coaching contributors and helping with pronunciation for those with borderline literacy \cite{emezue2025naijavoicesdatasetcultivatinglargescale}.

In the next section, we try to answer the more general question of the time required to create a speech annotated dataset by reviewing existing literature. Section \ref{section:3} presents our detailed analysis of one month of Bambara transcription, including the technical setup and the professional profile of the ten transcribers involved. In section 4, we conclude and discuss the results of this study in section 5.

\begin{figure*}[ht]
    \begin{center}
    \includegraphics[height=9cm, width=0.8\textwidth]{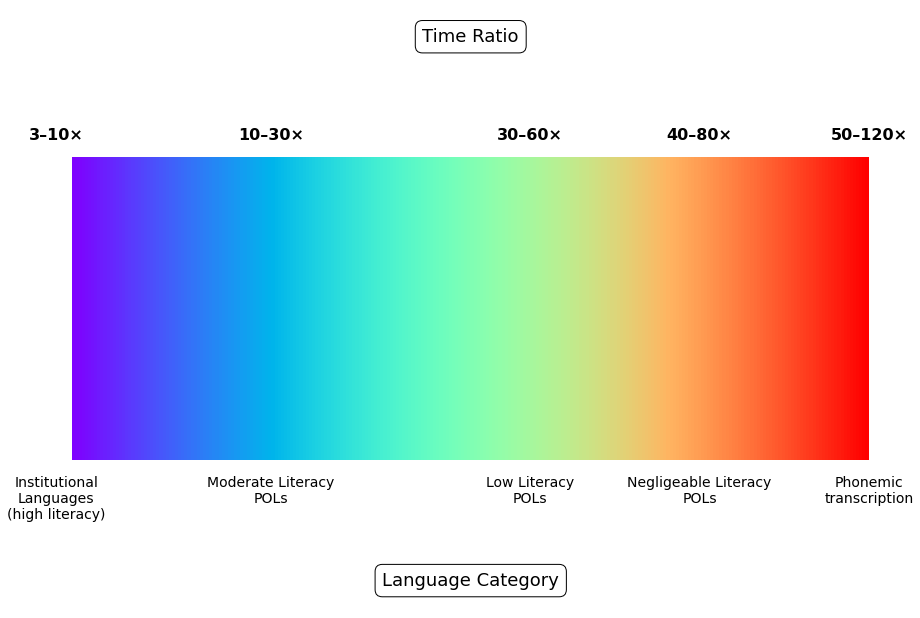}
    \caption{Transcription Effort Spectrum: Human Labor Multiplier per Language Category. The values on the top represent the time factor for transcribing an hour of audio against language community literacy labeled below the chart. We do a simple linear projection based on the data from published reports, our experiments and interpolation to make this visualization.}
    \label{fig1}
    \end{center}
\end{figure*}

\section{How much time to transcribe one hour of speech?}
\label{section:2}
While an abundance of data such as number of hours and tokens collected, average lengths and other classifications, is available for Institutional languages such as English, it is difficult to find a scientific article that clearly reports \textit{human-hours vs audio-hours} for transcription.

Professional transcription for English typically takes 3 to 6 hours per hour of audio (3-6x) and costs \$1.5 to \$3 per audio minute or \$90 to \$180 per audio hour in the US\footnote{GMR Transcription: \\https://www.gmrtranscription.com/blog/transcribing-an-hour-of-audio}\footnote{Notta: https://www.notta.ai/en/blog/how-long-does-it-take-to-transcribe-1-hour-of-audio}. Those seem to be the industry standards, the service may add additional fees for added complexities, for instance, a recording with over five people speaking can go up to 8 hours of labor per hour of recording.

The variance is already quite large for institutional languages depending on the exact conditions for the task. A 2024 SOTA review of Automatic Speech Recognition by Russell et al reports that even for English, manual transcription can take 14 times the length of the audio \cite{RUSSELL2024100163}.

In her analysis of transcription as an interpretive act rather than a straightforward technical task, Julia Bailey analyzes the various technical factors that can make transcription of 1 hour audio or video recording take anywhere between 3 to 10 hours. Although her analysis focuses specifically on the primary care and health sector, many of those factors such as capturing non-verbal and context relevant information in the transcript can easily transfer to any transcription task that is not purely verbatim \cite{Bailey2008transcribing}.

For POLs we have few sources for estimates. In 2019, Li et al presented SANTLR (Speech Annotation Toolkit for Low Resource Languages), which developed a three-stage ranking system for audio and text to optimize annotator's time and data variety. The toolkit prioritizes audio segments based on factors like duration, SNR, and phoneme overlap , while text is ranked using an LM perplexity score and edit distance overlap \cite{li2019santlrspeechannotationtoolkit}. While they do not clearly report transcription time during their experiment they do emphasize that in general annotators who had high computer literacy would perform those annotation tasks much faster and with more ease.

Prior work on a Bambara speech project, Jeli-ASR, estimated a team of 10 trained annotators took approximately 600 combined human hours to transcribe 10 hours of audio. This yields a \textbf{60x} ratio \cite{Diarra2022Griots}. The annotators were trained using the ELAN open source software for offline full-length audio transcription \cite{brugman-russel-2004-annotating}.

Though limited, the research indicates that the literacy level of a language community correlates to the transcription rate for voice recordings. Figure \ref{fig1} provides a visualization of this relationship. On the left side of the spectrum are Institutional Languages with \textit{high literacy} (\textit{eg}. English, French, Portuguese), followed by \textit{moderate}, \textit{low}, and \textit{negligible literacy} POLs, with languages without established writing systems on the right side of spectrum. Within language categories, there is a continuum relating to language resources specific to the language as well as other factors affecting transcription time. Thus, \textit{moderate literacy }languages such as Swahili and Hausa may be found on different points on the spectrum, as may be the case for \textit{low literacy } languages such as Bambara and Tamasheq and for \textit{negligible literacy} languages such as Bomu and Daza. Non-POL languages without established writing systems on the extreme right of the spectrum will require phonemic transcription, usually by trained linguists, which may take between 30 minutes to 2 hours for one minute of speech, depending on the transcriber’s familiarity with the language and the difficulty of the content \cite{adams-etal-2017-phonemic}.

\section{Analysis of the Experiment}
\label{section:3}
We conducted a study with ten Bambara transcribers to answer the main question of this article. In this section, we present the results and put them into perspective by defining the transcription setup and incorporating findings from an anonymous survey conducted with the transcribers. 

\subsection{Setup and Task Definition}
The transcription platform was built on Label Studio \cite{label2020studio}, using Google Cloud Storage (GCS) for data management and delivery. We had collected 612 hours of Bambara voice data from 500 speakers consisting of spontaneous speech in response to prompts spoken by \textit{facilitators} designed for cultural relevance. The audio files were segmented using Voice Activity Detection (VAD) \cite{li2019santlrspeechannotationtoolkit} and pre-transcribed using one of our pretrained ASR models, then the pre-labeled segments were imported in the labeling interface from GCS as individual and independent tasks.

The transcription task in our experiment is simplified as a first review task where the transcriber sequentially listens to an audio segment and corrects any mistakes in the automatic transcription. One task corresponds to one audio segment. The average segment duration was 2.2 seconds, ranging from 1 to 30 seconds. The transcribers demonstrated mastery of the process after 2 hours of training on the platform. 

\subsection{Transcribers Profile}
We present statistics on the profiles of the transcribers (education level, computer literacy, age, etc.)

Nine out of our 10 transcribers participated in the anonymous survey, 100\% of them have at least an undergraduate University degree. Three of them (33\%) studied Literature/Languages, 2 studied Social Sciences and one of them studied, respectively, Human Resources, Biological Sciences, Law, and Arts.

Six out of 9 annotators (66\%) said they have been studying Bambara for less than 3 years\footnote{This is the norm in Mali as French remains the language of instruction in schools, opportunities to formally study Bambara are limited to the university level.} and half of those have been developing Bambara literacy skills for less than a year. Two out of nine said they have been studying the language for 3-9 years and only one of them reported to have been studying Bambara for 12 years or more. Nine out of nine did it through a formal activity (certification/degree) and 100\% think they have either a good or excellent level of proficiency in the language (4-5 on a scale of 1 to 5).

Eight out of nine (88\%) said they have experience with office software, typing or other simple computer related tasks while one of them reported more advanced experience (programming or other specialized software manipulation). Five of them said this project was their first time doing an online annotation project. Nine out of nine reported that the transcription guidelines were perfectly understood and that Label Studio was easy to use, fast, and reliable.

We had 33\% of the transcribers outside of Bamako region, with one working from Equatorial Guinea. Finally, we had 3 transcribers, respectively, in each of the following age groups: 25-34, 35-44 and 45 or more.

\subsection{Performance Analysis}

\begin{table}
    \centering
    \begin{tabular}{lccc}
    \hline
     & Time (hours) & Avg segments & Efficiency \\
    \hline
    A1 & 10.1 & 622.6 & 61.6 \\
    A2 & 6.0  & 362.0 & 60.3 \\
    A3 & 7.0  & 322.0 & 46.0 \\
    A4 & 8.0  & 352.8 & 44.1 \\
    A5 & 8.2  & 376.4 & 45.9 \\
    A6 & 9.1  & 327.1 & 36.0 \\
    A7 & 10.1 & 353.0 & 35.0 \\
    A8 & 15.1 & 319.0 & 21.1 \\
    A9 & 5.0  & 329.6 & 65.9 \\
    A10 & 11.0 & 249.5 & 22.7 \\
    \hline
    \end{tabular}
    \caption{The Time column report the average time with the Labeling Interface open. Efficiency ratio express the number of segments per hour.}
    \label{tab1}
\end{table}

Over 25 working days, the team annotated 53.6 hours of voice data. This averages to 2.14 hours of annotated data per day. With a standard 8-hour workday for each of the 10 annotators, this yields a performance ratio of \textbf{36x}—meaning 36 hours of human labor were required for each hour of annotated audio. Table \ref{tab1} presents a more detailed statistical summary of the annotations. We can derive an estimate of the average audio duration per transcriber by multiplying the number of segments with the average segment duration.

The transcribers performed their work at home. Typical challenges that can impact work performance in a Malian home include frequent and prolonged power cuts, lack of an appropriate work space, and poor internet connectivity. While the Label Studio software enabled us to gather precise data on the real-time activity of each transcriber, we set up an observation lab to measure performance under supervision in ideal conditions. Three transcribers participated in the experiment, performing a full day of transcription work while senior project scientists observed them. We aimed to test the previous baseline against the performance achieved under controlled laboratory conditions and to gain additional insights into the transcription task for Bambara.

\begin{table}[ht!]
    \centering
    \begin{tabular}{lcccc}
    \hline
     & Time & Avg segments & Minutes & Ratio\\
    \hline
    A5 & 8  & 409 & 13.48 & 35.5x\\
    A7 & 6.5 & 371 & 13.33 & 29.6x\\
    A9 & 8  & 527 & 18.33 & 26.2x\\
    \hline
    \end{tabular}
    \caption{The Time column here represent the time at the office, including a 15 minutes lunch break and two 5 minutes prayer breaks. The third column reports the actual amount of minutes annotated during that work period, the ratio is thus expressed in hours this time}
    \label{tab2}
\end{table}

The transcribers worked for 8 hours except A7 who worked 6.5 hours. Table \ref{tab2} shows the results of that experiments and sets the laboratory baseline around \textbf{30x}.

Observation and discussion with the transcribers brought out the challenges they faced in navigating subtle dialectal variations and mentally mapping spoken phonemes to a written script that is not their primary mode of daily literacy. These linguistic complexities fundamentally distinguish this work from transcription in a \textit{high literacy} Institutional language or a \textit{moderate literacy} POL and are the primary driver of the high time-cost ratio. This cost seems to be fundamentally rooted in the inherent complexities of working with a \textit{low literacy} POL. 

\section{Conclusion}
This study establishes a critical empirical baseline: producing one hour of high-quality, corrected Bambara speech data requires \textbf{approximately 30 hours of human labor under ideal conditions}, with a 20\% overhead cost for performing the work under prevailing conditions. The cognitive demand, even for a simplified correction task are considerable.

Accurately budgeting for and scaling data creation for \textit{low literacy} POLs requires grounding our expectations in these linguistic realities, not just logistical ones. This 30x figure provides a scientifically-measured, experience-based starting point for such planning.

\section{Limitations}
A few factors warrant caution when interpreting the metrics and results presented in this study, while also highlighting directions for future work. Firstly, annotators noted that the transcription task became significantly more complex for longer audio segments (over 15 seconds) and recordings with poor audio quality, such as those affected by strong accents or background noise. Secondly, although all annotators had at least an undergraduate education and were proficient in Bambara, two-thirds had been studying its written form for less than three years, and about half were participating in their first online annotation project. Limited literacy in the writing system and modest computer skills likely contributed to the overall time expenditure. These constraints reflect the realities of working within a low-literacy language community. The results presented in this paper serve as a baseline taking into account these factors and the definition of the transcription task itself. Those can certainly be refined to improve the baseline and update the spectrum.

\section{Ethics Statement}
This study was conducted in full respect of the ethical principles of voluntary participation, informed consent, and data confidentiality. All Bambara transcribers were adult participants who provided informed consent prior to participation and were compensated fairly for their time and expertise. No personally identifiable or sensitive data were collected or shared. The research involved no vulnerable populations and posed no foreseeable risk to participants. Audio data used for transcription were sourced from consenting speakers under the RobotsMali AI4D Lab’s data collection framework, in compliance with institutional and national ethical guidelines for language data collection and processing.

\section{Acknowledgements}

The authors express their gratitude to the team of annotators whose meticulous work was the basis of this study. Our sincere thanks to Seni Tognine, Diakaridia Bengaly, Tahirou Mallé, Lanseni Mallé, Tidiane Koné, Aboubacar Traoré, Alassane Koné, Karounga Kanté, Boureima Traoré, and Benogo Fofana.

Finally, our thanks go to our technical team who developed and managed the annotation platform, the processing of the data, and the cloud storage, ensuring the smooth operation of the annotation platform despite the challenging local conditions.

\bibliographystyle{apalike} 
\bibliography{bib}

\end{document}